\documentclass[11pt,a4paper]{article}
\usepackage[hyperref]{emnlp2021}
\usepackage{times}
\usepackage{latexsym}
\usepackage{booktabs}
\usepackage{graphicx}
\usepackage{xspace}
\usepackage{float}

\usepackage{microtype}

\newcommand*\samethanks[1][\value{footnote}]{\footnotemark[#1]}

\title{Faithful and Plausible Explanations of Medical Code Predictions}

\author{Zach Wood-Doughty\thanks{\hspace{1mm} Equal contribution}, Isabel Cachola\samethanks[1], Mark Dredze \\
  Johns Hopkins University \\
  Baltimore, MD, 21211 \\
  \texttt{zach@cs.jhu.edu, icachola@cs.jhu.edu, mdredze@cs.jhu.edu} \\ }

\date{}

\begin{document}
\newcommand\isabel[1]{{\color{purple}\{\textit{#1}\}$_{Isabel}$}}

\newcommand\todoit[1]{{\color{red}\{TODO: \textit{#1}\}}}
\newcommand\todo{{\color{red}{TODO}}\xspace}
\newcommand\todocite{{\color{red}{CITE}}\xspace}
\maketitle
\begin{abstract}
Machine learning models that offer excellent predictive performance often lack the interpretability necessary to support integrated human machine decision-making.  In clinical medicine and other high-risk settings, domain experts may be unwilling to trust model
predictions without explanations. Work in explainable AI must balance competing objectives along two different axes: 1) Explanations must balance
\emph{faithfulness} to the model's decision-making with their \emph{plausibility} to
a domain expert. 2) Domain experts desire \emph{local} explanations of
individual predictions and \emph{global} explanations of behavior in aggregate. 

We propose to train a proxy model that mimics the behavior of the trained model and provides fine-grained control over these trade-offs.
We evaluate our approach on the task of assigning ICD codes to clinical notes to demonstrate that explanations from the proxy model are faithful and replicate the trained model behavior.
\end{abstract}

\section{Introduction}
Machine learning (ML) methods have demonstrated predictive success in medical settings, leading to discussions of how ML systems can augment clinical decision-making \cite{caruana2015intelligible}. However, a prerequisite to clinical integration is the ability for healthcare professionals to understand the justifications for model decisions. Clinicians often disagree on the proper course of care, and share their justifications as a means of agreeing on a treatment plan. Explainable Artificial Intelligence can enable models to provide the explanations needed for them to be integrated into this process. A significant challenge to explanations comes from the complexity of modern AI models, which rely on complex deep neural networks with millions or billions of parameters.

Similar concerns over model explanations across domains have inspired a whole field of interpretable ML. Work in this area has considered two goals: faithfulness -- explanations that accurately convey the decision making justifications of the model -- and plausibility -- explanations that make sense to domain experts. Balancing these goals can be challenging, as more faithful explanations that accurately convey the reasoning of complex AI systems may be implausible to a domain expert, and vice versa. An additional trade off must consider using sophisticated versus interpretable methods. The sophisticated methods may yield the best performance on a task, but be least able to provide explanations.

We propose to use a proxy model to disentangle the conflicting goals of sophisticated methods with high accuracy and interpretability. We train a fundamentally interpretable model -- such as logistic regression -- on the \emph{predictions} of the target ML model, so that the behavior of the proxy model mimics the target model's behavior, rather than independently modeling the target task. We then rely on the interpretable proxy model to create explanations, allowing the trained model to use sophisticated methods to achieve high accuracy. %Furthermore, by relying on a separate proxy model we can explicitly tune it to trade off between faithfulness and plausibility, as well as local versus global explanations, all while not imposing any constraints on the high accuracy trained model. 
Validation of this approach requires answering two questions: 1) Is the proxy faithful to the workings of the trained model? and 2) Are the produced explanations of high quality to domain experts?
 
We demonstrate our approach on the task of medical code prediction, following \citet{mullenbach2018explainable}. While ML methods have achieved predictive success on various versions of ICD clinical code prediction, the best-performing methods have been neural networks that are notoriously difficult to interpret. \citet{mullenbach2018explainable} introduced DR-CAML, a method designed to produce explainable predictions and they evaluated predictions with annotations from clinicians. Their work showed that DR-CAML produces more plausible explanations than reasonable baselines.

We reproduce their work and compare it to our approach of a proxy model.%, allowing us to explicitly trade off between faithful and plausible explanations. 
We use a linear logistic regression proxy model that learns to mimic the behavior of the trained DR-CAML model. We show that the proxy model is faithful to the original model and produces plausible explanations, as measured on the clinical annotations of generated explanations. 
% We also conduct a detailed analysis of the behavior of the proxy model and the resulting explanations, showing how our approach works in practice. 
We release our code and work on reproducing~\citet{mullenbach2018explainable}.

\section{Background}

\subsection{Interpretable ML}

Interpretable machine learning is a major area within the growing field of Explainable AI~\cite{doshi2017towards}. We present an overview of major themes in the literature, and direct the reader to recent surveys for more details~\cite{doshi2017towards,guidotti2018survey,gilpin2018explaining}.

Past work distinguishes between “transparent” or “inherently interpretable” models that offer their own explanations, and “post-hoc” methods that produce explanations for a separately-trained model.
Methods such as logistic regression are often considered transparent or inherently interpretable, because their simplicity allows a domain expert to understand how a change in input would produce a different output~\cite{guidotti2018survey}.
However, even simple models can prove difficult to interpret in certain settings, such as when the model's features are complex~\cite{lipton2018mythos}.
LIME is an example of a post-hoc method~\cite{ribeiro2016should}; given a trained model of arbitrary complexity it produces explanations for individual predictions.
The trade-off inherent and post-hoc methods is that inherently-interpretable methods are often limited in model complexity.
Deep neural networks, for example, often demonstrate better performance but are not inherently interpretable~\cite{feng2018pathologies},
and typically rely upon post-hoc methods to derive explanations~\cite{guidotti2018survey}.

\citet{lipton2018mythos} critiques the idea of ``inherent'' interpretability and argues that methods that are intended to be transparently understood should pursue several traits.
These include simulatability, or whether a human can reasonably work through each step of the model's calculations to understand how a prediction is made; decomposibility, or whether each parameter of the model can be intuitively understood on its own; and algorithmic transparency, or whether the model belongs to a class with known theoretical behaviors.
\citet{lou2012intelligible} highlights linear and additive models as particularly decomposible (or intelligible) classes of models, because ``users can understand the contribution of individual features in the model.''
Our proposed approach will use a linear model trained on bag-of-word features to provide a simulatible, decomposible, and transparent method.

Interpretability methods are also distinguished by the form and quality of the explanations it produces. Two primary desiderata for explanations of ML systems are “faithfulness” and “plausibility.”\footnote{Faithfulness is also referred to as validity or completeness; plausibility is alternatively referred to as persuasiveness~\cite{herman2017promise} See \citet{jacovi2020towards} for a longer discussion of alternate terminology.}
A faithful method accurately describes the true machinery of the model’s prediction, while a plausible model produces explanations that can be interpreted by a human expert~\cite{jacovi2020towards}.
A method could be faithful but not plausible, if it accurately explains a model’s predictions but does so in terms of high-dimensional feature vectors that a human cannot interpret.
Similarly, a method could be plausible but not faithful if it produces concise natural language summaries that are unrelated to the calculations that produce the model’s predictions.
Methods should attempt to achieve both goals, but there is a trade-off between the two; explanations typically cannot be both perfectly descriptive and concise.
Plausibility, unlike faithfulness, necessarily requires an evaluation based on human perception~\cite{herman2017promise}. A strength of our proposed method is that it is designed for plausibility and transparency, but optimized for faithfulness.
% A strength of our proposed method is that it can explicitly navigate this trade-off between faithfulness and plausibility.

\subsection{Explainable prediction of medical codes}

Our work closely follows that of \citet{mullenbach2018explainable}. We use the same dataset of clinical texts and associated medical codes (described in \S~\ref{sec:faithfulness}) and compare against their method, called Description-Regularized Convolutional Attention for Multi-Label classification (DR-CAML).
DR-CAML is a neural model that seeks to produce its own faithful explanations using a per-label attention mechanism that highlights n-grams in the input text that were correlated with the model's predictions.
Because DR-CAML has over six million learned parameters, it does not fulfill simulatability or decomposability; a single parameter cannot be understood in any intuitive way.
However, the attention mechanism allows for some insight into the model's decision-making, as it indicates which regions of the input text were given more weight in the final classification.

DR-CAML’s use of attention to produce explanations has sparked discussion.
\citet{jain2019attention} showed that attention mechanisms can provide misleading explanations that are not faithful to the model’s true reasoning.
\citet{wiegreffe2019attention} has argued that while the explanations produced by attention may not always be faithful, they are often plausible.
This discussion has continued in the interpretable ML literature, with methods demonstrating how attention mechanisms can be useful or deceptive~\cite{zhong2019fine,grimsley2020attention,jain2020learning,pruthi2020learning}. Creating models that are both faithful and plausible remains a challenge.
% A remaining challenge is that while we may be able to measure the faithfulness or plausibility of an explanation, rarely can a method be easily tuned to produce more faithful or more plausible explanations.

\begin{figure*}[!t]
    \centering
    \includegraphics[width=0.7\textwidth]{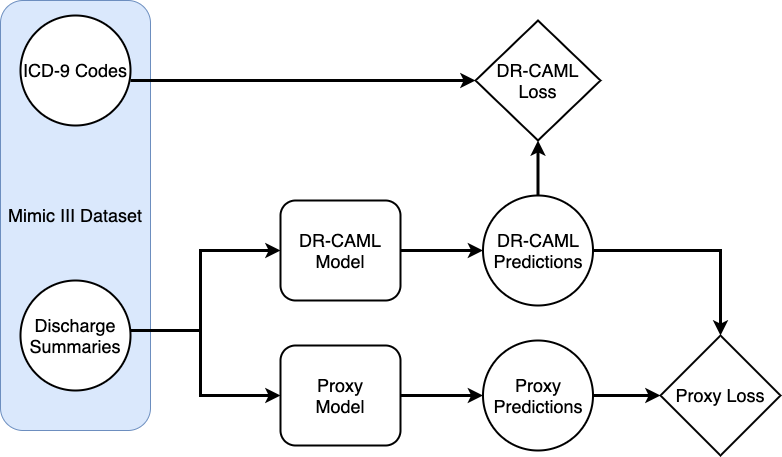}
    \caption{Relationship between trained DR-CAML model and proxy model. The proxy model is trained to predict DR-CAML's outputs, rather than the true ICD-9 codes. This optimizes the proxy model for faithfulness.}
    \label{fig:proxy_outline}
\end{figure*}
\section{Methods}

% Our proposed method seeks to provide the flexibility to balance faithfulness and plausibility.
Our proposed method is post-hoc and seeks to to balance faithfulness and plausibility.
We assume that we have a trained model with good predictive performance, but low interpretability.
Given this trained model and a dataset on which it can be applied, we
train a {\it proxy model} that takes the same input from the dataset, but uses the trained model's predictions as its labels. In other words, given the dataset's input, the proxy model predicts the outputs of the uninterpretable model.
Figure~\ref{fig:proxy_outline} gives a visual representation of the proxy model setup.
For the medical code classification task, the original model (DR-CAML) is trained on the text of discharge summaries and produces a probability for each of the 8,922 possible medical codes.
We apply DR-CAML to the texts in MIMIC III and save its continuous-valued probabilities as the labels for our proxy model.
By training the proxy model to output predictions similar to that of the existing model, we are optimizing it to be faithful by design.

We also desire that the proxy model produces plausible explanations and fulfills the criteria from \citet{lipton2018mythos}: simulatibility, decomposibility, and algorithmic transparency.
To do so, we restrict our proxy model to a class of models that fulfills these desiderata.
The fundamental trade-off here is that if we restrict our model class too much, the proxy will be unfaithful, unable to mimic the behavior of the trained model.
But if we allow for a proxy model that is too complex, it may not provide plausible or otherwise desirable explanations.
The choice of proxy model requires some consideration of the particular domain, as feature preprocessing and similar details may affect what explanations are possible.

For the specific task of medical code prediction, we will use a linear regression model trained on a bag-of-words representation of the clinical texts.
We train 8,922 proxy models, one for each medical code in the dataset's labels.
We implement our method using the linear {\tt SGDRegressor} model from {\tt sklearn}~\cite{pedregosa2011scikit},
and apply a log transform to the model’s probability outputs and train the proxy to minimize squared loss.
We release the code for training and evaluating our method.\footnote{\href{https://github.com/isabelcachola/mimic-proxy}{github.com/isabelcachola/mimic-proxy}}

Our approach is similar to LIME~\cite{ribeiro2016should} in that it learns a simple (linear) model to explain a pretrained model.
However, whereas LIME learns a linear model to post-hoc explain a single prediction, our linear model is trained to predict and explain the entire dataset of predictions.
This has several consequences. Unlike LIME, we do not require sampling perturbed inputs that do not exist in the training data, which can produce contrasts which are misleading or unintuitive~\cite{mittelstadt2019explaining}.
\citet{slack2020fooling} has shown that LIME can be fooled into providing innocuous explanations for models that demonstrate racist or sexist behavior by exploiting its reliance on perturbations. 
It also means that our proxy model is given a more difficult task than a LIME model -- it may be that a given proxy model is insufficiently flexible to model the complexity of the pretrained model, in which case we can measure this failure in terms of our faithfulness evaluation (see \S~\ref{sec:faithfulness}).
Because LIME trains a model linear only in the neighborhood of a given instance, its feature importance scores are difficult to aggregate across a dataset, making extrapolation difficult~\cite{van2019global}.
When our proxy model is faithful to the trained model, our approach gives us explanations that we can expect to apply to future predictions.
If the proxy model demonstrates sufficient empirical performance, a domain expert may even prefer to use it in place of the original trained model, an option that LIME explanation models do not support.

By applying our proxy model method to the DR-CAML model from \citet{mullenbach2018explainable}, we enable an evaluation of both faithfulness and plausibility.
We will evaluate whether our model is faithful by seeing how closely its outputs match the predictions of DR-CAML.
Because DR-CAML was designed to be interpretable using its attention mechanism, we can compare its explanations against those produced by our proxy.
In the next two sections, we introduce our evaluation for the proxy model's faithfulness to the DR-CAML model and the plausibility of its explanations.

\begin{table*}[ht]
\centering
\begin{tabular}{c c ccc c cccc}
 & & \multicolumn{3}{c}{Regression} & & \multicolumn{4}{c}{Classification} \\
& &&& && \multicolumn{2}{c}{AUC} & \multicolumn{2}{c}{F1} \\
Model && Spearman & Pearson & Kendall && Macro & Micro & Macro & Micro \\
\cmidrule(r){1-1} \cmidrule(lr){3-5} \cmidrule(l){7-10}
Logistic && 0.036    & -0.195    & -0.135 && 0.734 & 0.936 & 0.012 & 0.353 \\
Proxy    && 0.498    & 0.794     & 0.608  && 0.980 & 0.995 & 0.052 & 0.416 \\
\end{tabular}
\caption{
Comparison of the logistic baseline and the proxy model to the DR-CAML predictions.
For the F1 evaluation, we threshold the unnormalized proxy outputs at 0.5.
The logistic model was trained to predict the ICD codes; the proxy model to predict DR-CAML's predictions.
As expected, the proxy model dramatically outperforms the logistic baseline in terms of faithfulness to the DR-CAML model.
\label{tbl:faithfulness}
}
\end{table*}

\begin{table}[!t]
\centering
\begin{tabular}{r ccc}
& Logistic & Proxy & DR-CAML      \\
\midrule
Macro AUC & 0.596 & 0.901 & 0.906 \\
Micro AUC & 0.889 & 0.967 & 0.972 \\[1em]
Macro F1  & 0.033 & 0.142 & 0.224 \\
Micro F1  & 0.278 & 0.326 & 0.536 \\[1em]
Prec @ 8   & 0.547 & 0.483 & 0.701 \\
Prec @ 15  & 0.413 & 0.407 & 0.548 \\
\bottomrule
\end{tabular}
\caption{
Comparison of the logistic baseline, the proxy model, and DR-CAML to true ICD labels.
Although the logistic model was trained for this specific task and the proxy model was not, the proxy model outperforms the baseline in terms of AUC and F1.
The proxy model's outputs are unnormalized, which partially explains the gap between its F1 scores, which are computed with a threshold of 0.5, and its AUC scores, which are invariant to normalization.
This lack of normalization may also explain the poor performance on precision metrics, as the proxy predicts each code independently.
}
\label{tbl:true_labels}
\end{table}

\section{Faithfulness evaluation} \label{sec:faithfulness}

The MIMIC-III dataset contains anonymized English-language ICU patient records, including physiological measurements and clinical notes~\cite{johnson2016mimic}.
Following \citet{mullenbach2018explainable}, we focus on discharge summaries which describe a patient's visit and are annotated with ICD-9 codes.
There are 8,922 different ICD-9 codes that describe procedures and diagnoses that occurred during a patient's stay. The manual assignment of these codes to patient records are required by most U.S. healthcare payers~\cite{topaz2013icd}.
We duplicate the experimental setup of \citet{mullenbach2018explainable} which uses the text of the discharge summaries as input to the DR-CAML model, which then is trained to predict all ICD-9 codes associated with that document.
After applying their pre-processing code to tokenize the text, the dataset contains 47,724 discharge summaries which are divided into training, validation, and test splits.

Our proxy model is the combination of 8,922 linear regression models trained to predict DR-CAML's log probability for each ICD-9 code.
After a brief grid search on the validation set, we chose to apply L1 regularization with $\alpha=0.0001$ for each regression.
To establish that this collection of linear regressions is faithful to the trained DR-CAML model, we want to show that it makes similar predictions across all ICD-9 codes on held-out data.
Recall from Figure~\ref{fig:proxy_outline} that the proxy is trained not to predict the true ICD-9 codes but to output the same label probabilities as DR-CAML. In fact, the proxy model never sees the true ICD-9 codes.
We evaluate faithfulness by comparing the outputs of DR-CAML and the proxy model on the held-out test set. If the two systems produced identical outputs on held-out data, we would say that the proxy was perfectly faithful.
We make this comparison in three different ways -- first using regression metrics that compare the continuous outputs of the two models, then using classification metrics with binarized DR-CAML predictions, and finally by using the proxy model's outputs perform when treated as predictions for the true ICD-9 codes.
For all these comparisons, we use a logistic regression baseline that is trained to directly predict the ICD-9 codes without knowledge of DR-CAML's predictions.
We would expect the logistic baseline's predictions to still be slightly correlated with those of DR-CAML, but not to be a faithful proxy itself.

Our first evaluation uses regression metrics that assess the correlation between the proxy’s predictions and DR-CAML’s predicted probabilities.
We use Spearman and Pearson correlation coefficients and the non-parametric Kendall Tau rank correlation. These metrics range from -1 to 1, where perfectly faithful predictions would have correlations of 1. These Regression results are on the left side of Table~\ref{tbl:faithfulness}. 

Our second evaluation treats DR-CAML’s predictions as binary labels based on whether they exceed the threshold used by \citet{mullenbach2018explainable} to compute F1 scores.
We then evaluate the faithfulness of our proxy model by treating its outputs as unnormalized probabilities and using classification metrics such as F1 score.
These metrics range from 0 to 1, where perfectly faithful predictions would have 1.0 AUC and F1 scores.
The proxy model is considered faithful if it correctly predicts whether DR-CAML will make a binary prediction.
We again use the logistic regression baseline.
These Classification results are on the right side of Table~\ref{tbl:faithfulness}. 

Finally, we use the proxy model’s predictions to predict the ground-truth ICD code labels and compare its predictive performance against that of DR-CAML in Table~\ref{tbl:true_labels}.
While the proxy model was not trained using these labels, we can use its predictions as unnormalized probabilities for these codes.
By comparing against the logistic regression baseline (a linear model of equal complexity), we can see whether our training setup allows the proxy model to learn a better predictor.

Our results show that the proxy model is quite faithful to the DR-CAML model.
Compared to the logistic regression baseline, the proxy model is dramatically better on all metrics in Table~\ref{tbl:faithfulness}.
Combining the results from Tables~\ref{tbl:faithfulness} and~\ref{tbl:true_labels} we can see that on AUC metrics, the proxy model is closer to the DR-CAML predictions than DR-CAML is to the ground-truth labels.
The proxy model also outperforms the logistic regression baseline in the classification metrics (AUC and F1), indicating that the proxy model is more faithful to the DR-CAML predictions.
In Table~\ref{tbl:true_labels}, we see a large gap between its performance on the AUC metrics and the F1 and precision metrics. This is likely because the outputs of the proxy model are not normalized to be valid probabilities, and AUC is invariant to normalization while and F1 and precision metrics are not.

\citet{rudin2019stop} critiques post-hoc methods in general, arguing that ``if we cannot know for certain whether our [post-hoc] explanation is [faithful], we cannot know whether to trust either the explanation or the original model.''
Because no post-hoc method can ever be perfectly faithful to an original model, we believe our approach to explicitly measuring faithfulness provides a useful approach for understanding whether the proxy is ``faithful enough'' for a given application.
It also allows for a prediction-specific analysis -- if we wish to use the proxy model to explain a high-stakes prediction made by DR-CAML, we can first check to see whether the two models agree upon that specific prediction.
In applications where explainability is paramount over accuracy, our proxy model could be used as a more interpretable replacement for a high-performing black-box model.
In such a case, the evaluation of faithfulness in Table~\ref{tbl:faithfulness} could be ignored in favor of the comparison in Table~\ref{tbl:true_labels}.
We leave for future work the question of whether the proxy model could be tuned or its predictions could be normalized to improve its performance on F1 or other metrics comparing it to the ground truth labels.

\section{Plausibility Evaluation} \label{sec:plausibility}

Explanations are considered plausible if they can be reasoned about by a human user.
Thus, evaluating plausibility is typically more difficult than faithfulness, because it requires input from human annotators~\cite{herman2017promise}. 
An explanation that is plausible to a domain expert may not be plausible to a layperson.
\citet{mullenbach2018explainable} evaluated the plausibility of CAML’s explanations by collecting annotations from a clinician.
\citet{wiegreffe2019attention} has argued that the attention mechanism of CAML and DR-CAML generally provide plausible explanations, even if they at times not be faithful to the model's internal decision-making.
For each model they considered in their paper, they extracted an explanation in the format of a 14-token subsequence taken from the discharge summary.
The clinician read all four explanations and the correspond ICD code and rated each explanation as either ``informative'' or not.
They found that CAML was rated as slightly more informative than their logistic regression and CNN baselines.

\begin{table}[!t]
    \centering
    \begin{tabular}{c c c}
    \toprule
    Model       & Score & Interval   \\
    \midrule
    Logistic    & 35    & (31, 50)  \\
    Cosine      & 38    & (33, 52)  \\
    CAML        & 44    & (33, 52)  \\
    DR-CAML     & 48    & (34, 52)  \\
    Proxy       & 65    & (39, 59)  \\
    \bottomrule
    \end{tabular}
\caption{
Plausibility evaluation using classifier annotations.
The Score column is out of 99; the proxy model is significantly better than DR-CAML according to a McNemar test with $p < 0.01$, and better than all other models with $p < 0.001$.
The Interval column shows a 95\% bootstrap interval using annotations sampled from their output probabilities.
Because the classifier is not calibrated, these sampled intervals heavily skew towards lower scores.
}
    \label{tbl:plausible}
\end{table}

The format of \citet{mullenbach2018explainable}'s plausibility evaluation does not easily lend itself to replication.
While the authors shared their annotations with us, the loss of important metadata made it difficult to directly reproduce their analysis.
Additionally, as the clinical annotator considered explanations in a comparative setting, it is not trivial to compare our proxy model against these same annotations.
Our goal was to replicate this evaluation by using a simple classifier to predict synthetic labels as to whether the clinical domain expert {\it would have} labeled our models' explanations as plausible.
Using BioWordVec embeddings released by \citet{zhang2019biowordvec}, the text of the ICD-9 code description, and the 14-gram explanation produced by each model from \citet{mullenbach2018explainable}, we train a classifier that predicts whether an explanation would have received an ``informative'' or ``highly informative'' label.
This annotation classifier achieves an accuracy of 67.2\% and an AUC score of .726 on held-out explanations, indicating it is a useful but noisy stand-in for the clinician.
The full training and evaluation details of this classifier are in Appendix~\ref{subsec:classifier}.

To conduct our plausibility evaluation, we first use or reimplement the baseline methods from \citet{mullenbach2018explainable}.
Each model, including the proxy, produces a 14-token explanation from the discharge summary by first finding the 4-gram with the largest {\it average feature importance} and then including five tokens on either side of the 4-gram.
The logistic regression baseline is the same as considered in \S~\ref{sec:faithfulness}, where feature importance is computed simply using the coefficients of the logistic model.
The proxy model's explanations are computed in the same manner, finding the 4-gram with the largest average coefficient weights.
The CAML and DR-CAML models extract explanations using their attention mechanism, as implemented in the code released by ~\citet{mullenbach2018explainable}.
Finally, we reimplement their cosine similarity baseline, which simply finds the 4-gram in the discharge summary with the highest cosine similarity to the words in the ICD-9 code description.

We extract the model's explanations for the same\footnote{Using the 99 (of 100) discharge summaries that could be uniquely identified. See Appendix~\ref{sec:appendix} for details.}
discharge summaries as were evaluated by \citet{mullenbach2018explainable}.
For each explanation, we use the annotation classifier described above to predict the probability that each explanation would have been labeled as informative.
If we set the classifier threshold such that 42\% of explanations are rated as informative (matching the proportion from the original annotations), we get the results in the Scores column of Table~\ref{tbl:plausible}.
We see that the proxy model significantly (according to a McNemar test) outperforms all other models.
We then highlight the uncertainty introduced by relying on our imperfect annotation classifier.
Rather than thresholding the outputs of the annotation classifier, we instead use the probabilities it outputs and sample a set of informative labels for each explanation.
We sample 1000 such sets of labels and report the 95\% confidence interval for each model's score in the Interval column of Table~\ref{tbl:plausible}.
Accounting for this uncertainty dramatically reduces the differences between the methods.
However, because and 95\% of all classified plausibility probabilities are between 23.6 and 61.2, this sampling method makes it unlikely that any bootstrap interval would reach the proxy model's thresholded score of 65.
If, instead of resampling annotations, we vary the threshold for a plausibility label, the gap between the proxy model and other methods widens.
If we set the threshold so only 10\% of explanations are classified as informative, the proxy model achieves a score of 34, whereas the next best model (CAML) achieves a score of only 6.
Despite the inherent uncertainty involved in extrapolating plausible scores from a fixed set of clinical annotations, our evaluations suggest that the proxy model produces explanations that are at least as plausible as those of the DR-CAML model on which it is trained.

\section{Discussion} \label{sec:discussion}

We have introduced a method for post-hoc explanations that is designed to be interpretable and plausible while maintaining faithfulness to the trained model.
By constraining the proxy to a class of models that is decomposible, simulatible, and algorithmically transparent, our optimization for faithfulness gives us a clear way to evaluate several dimensions of interpretability. Furthermore, our proxy model only has 50K parameters, compared to CAML's 6 million. 
A key benefit of our method is its simplicity and wide applicability.
Even for a proprietary trained model for which the learned parameters are unknown, a proxy can be trained as long as we have a dataset that includes the trained model's predictions.
Our approach has the additional benefit of producing a standalone proxy model that can provide global, rather than local, explanations.
Depending on the gap in predictive performance between the proxy and original model, a skeptic of post-hoc methods (e.g. \citet{rudin2019stop}) might prefer to discard the original model altogether and simply make predictions using the proxy model, for which its explanations are faithful by design.

The present work has several limitations that are left to be addressed in future work.
Though the task of medical code prediction has important implications and has been widely studied in interpretability research, we only consider this single task on a single English-language dataset.
We believe this proxy model approach is generally applicable as a post-hoc interpretability method for arbitrary models, but this must be further studied on new datasets and different trained models.
It is possible that some in future domains, trained models might be more difficult to mimic than DR-CAML. If so, the application may require a trade-off between a less restrictive proxy model class or a lower faithfulness of the proxy model.

Our evaluation is also limited in that it only considers a single form of explanations: n-grams extracted via feature importances or attention weights.
Recent work has explored alternate formulations for a quality explanation~\cite{barocas2020hidden}; some formulations may be more or less accommodating of our proxy model method.
Our plausibility evaluations rely heavily on a single set of expert annotations from which we extrapolate using a classifier.
To demonstrate that our method can reliably provide both plausible and faithful explanations, additional evaluations must collect new plausibility annotations or build off of existing resources \cite{deyoung2020eraser}.

As the ML community continues to explore new directions for interpretable methods, definitions of desiderata may continue to evolve.
Such criteria will always depend on the domain experts who turn to an ML method for decision support.
Interpretable ML methods should clearly define how they expect to satisfy a criterion such as faithfulness or plausibility.
By designing for plausibility and transparency and optimizing for faithfulness, our proposed method is broadly applicable.
We release our code to enable future extensions of our work.

\section*{Acknowledgements}

We thank Sarah Wiegreffe and Jacob Eisenstein for their help and plausibility annotations.

\clearpage

\bibliography{anthology,acl2020}
\bibliographystyle{acl_natbib}

\clearpage

\appendix

% \begin{minipage}{\linewidth}
\begin{table*}[!h]
\centering
\begin{tabular}{lll ll ll}
 & \multicolumn{2}{c}{AUC} & \multicolumn{2}{c}{F1} & \multicolumn{2}{c}{P@n} \\
 \cmidrule{2-7} 
 & Macro & Micro & Macro & Micro & 8 & 15 \\ \hline
\multicolumn{1}{l}{\citet{mullenbach2018explainable}} & 0.895 & 0.986 & 0.088 & 0.539 & 0.709 & 0.561 \\
\multicolumn{1}{l}{\citet{wiegreffe2019clinical}} & 0.889 & 0.985 & 0.080 & 0.542 & 0.712 & 0.562 \\
\multicolumn{1}{l}{Ours (using released weights)} & 0.892 & 0.978 & 0.090 & 0.298 & 0.636 & 0.471 \\
\multicolumn{1}{l}{Ours (retrained)} & 0.628 & 0.884 & 0.001 & 0.024 & 0.042 & 0.027
\end{tabular}
\caption{
Published predictive performance of CAML and our replicated results.
Our experiments throughout the paper use the model with the released weights, which is closest to the published numbers (despite Micro F1).
\label{tbl:caml_reproduction}
}
\end{table*}
% \end{minipage}

\section{(Re-)implementation details} \label{sec:appendix}

\subsection{Reproducing CAML predictive performance}

The trained DR-CAML model released by \citet{mullenbach2018explainable} produced predictions that matched the published F1 and ROC scores.
We were unable to precisely replicate the outputs of the CAML model.
Table~\ref{tbl:caml_reproduction} shows the scores published by \citet{mullenbach2018explainable} as well as those for a CAML reimplementation done by \citet{wiegreffe2019clinical}.
We include the scores we observe using the model weights released on GitHub as well as the scores for a model we retrained from scratch.  
We use the released model instead of the retrained model as its performance is much closer to the published numbers.

\begin{table*}[!ht]
\centering
\begin{tabular}{l c c c c c}
Model       & Theirs & Ours & E1 & E2 & Full \\
\midrule
CAML        & 46   & 54    & 47     & 43    & 44    \\
DR-CAML     & --   & --    & 45     & 44    & 48    \\
Cosine      & 48   & 48    & 41     & 40    & 38    \\
Logistic    & 41   & 43    & 47     & 49    & 35    \\
CNN         & 36   & 46    & 51     & 47    & --    \\
\bottomrule
\end{tabular}
\caption{
Plausibility evaluations and comparison to \citet{mullenbach2018explainable}.
The Theirs column shows the published numbers; Ours shows our best attempt at matching the clinical evaluation to the trained models.
While the numbers change dramatically, the ordering only changes by two swaps.
E1 and E2 show the results with predicted plausibility labels under two evaluation settings.
Full duplicates the results from Table~\ref{tbl:plausible} for comparison.
While we were able to retrain the CNN model for the E1 and E2 evaluations, the implementation was unable to run on the GPUs we had access to.
}
\label{tbl:plausibility_reproduction}
\end{table*}

\subsection{Reproducing plausibility scores} \label{subsec:plausibility_reproduction}

The clinical plausibility annotations provided to us by the authors of ~\citet{mullenbach2018explainable} contains the text explanations and their corresponding annotations, but was missing the crucial metadata of which models produced which explanations.
The metadata also did not indicate from which specific discharge summary the texts were derived; while the text explanations were uniquely identifying for all but one of the 100 examples.
For that one example, because some patients had multiple documents sometimes containing duplicated segments of text, there were three discharge summaries from which the explanations could have been drawn.
We thus excluded this example from our analyses.
To replicate their analysis the best we could, we retrained or reimplemented their logistic regression, vanilla CNN, and cosine similarity methods.
We then looked at the attention or feature importance weights for each trained model and the text explanations that had been annotated, and assigned each model the text explanation for which it provided the highest weight.
This assignment did not perfectly align with past work: there were six cases (out of 99) where a text explanation was "chosen" by more models than times it appeared as an option.
Ignoring that issue and then simply aggregating the Informative and Highly Informative clinician annotations, we obtained the plausibility scores in the Ours column of Table~\ref{tbl:plausibility_reproduction}.
The Theirs column shows the published numbers from \citet{mullenbach2018explainable}.
While the numbers change substantially, the ordering is relatively stable with only two swaps: CAML and Cosine, and Logistic and CNN.
The other columns of the table are described below.

\subsection{Plausibility annotation classifier} \label{subsec:classifier}

To evaluate the plausibility of our proxy model's explanations,
we trained a classifier to predict whether an explanation would have been labeled as plausible by the clinical domain expert.
We treat this as a binary classification task by grouping the "Informative" and "Highly Informative" annotations as a single "plausible" label.
Conscious of the fact that we have only 99 examples with four text explanations each, we use two approaches with which to train and evaluate our classifier.
The first used leave-one-out cross validation at the example level, such that the classifier was trained on 98 examples at a time and then evaluated on the remaining one.
We refer to this evaluation as "E1" in Table~\ref{tbl:plausibility_reproduction}.
The second also used leave-on-out cross validation but at the explanation level; we held out a single text explanation, trained on all other explanations across all examples, and then evaluated on the held-out explanation.
When an explanation appeared more than once in a single example, we made sure to remove its duplicates from the training data for predicting that explanation.
We refer to this evaluation as "E2" in Table~\ref{tbl:plausibility_reproduction}.

The trained model is a simple logistic regression classifier trained on a fastText embedding of both the explanation and the target ICD-9 code description.
Using the BioWordVec embeddings released by \citet{zhang2019biowordvec}, we embed each both the explanation and code description into a 200-dimensional vector, concatenate the two vectors, and pass it to the logistic regression.
In the E1 evaluation, the model achieves an accuracy of 60.6\% and an ROC AUC score of .640. 
In the E2 evaluation, that increases to an accuracy of 67.2\% and an AUC score of .726,
indicating that the additional within-example explanations substantially help the classifier.

When using these classifiers to label the explanations generated by each model instead of the plausibility scores derived in \S~\ref{subsec:plausibility_reproduction}, we get the results shown in columns E1 and E2 of Table~\ref{tbl:plausibility_reproduction}.

Finally, we retrain our final classifier on all the explanations, leaving none held out.
Rather than using our classifier to evaluate the explanations that were actually shown to the clinician, we instead use our (re-)implementation of the four models to extract an explanation from each of the 99 discharge summaries.
These explanations thus may or may not appear in the training data for the classifier.
For this evaluation we are not worried about the classifier overfitting, as the classifier functions as a direct replacement for the clinician who produced the training data.
The results of this analysis are the numbers shown in Table~\ref{tbl:plausible} in \S~\ref{sec:plausibility}, reproduced in Table~\ref{tbl:plausibility_reproduction} in the "Full" column.
The Logistic model does much worse on the Full evaluation than in either E1 or E2.
This may be because the explanations selected by the trained model were worse than those selected by the model which was used for the original clinical evaluation.
We were unable to extract explanations from the retrained CNN model, which means we were unable to evaluate it for the Full setting.

\end{document}